\documentclass[11pt,a4paper]{article}
\usepackage[utf8]{inputenc}
\usepackage{microtype}
\usepackage[hyperref]{naaclhlt2019}
\usepackage{epstopdf}
\usepackage{times}
\usepackage{latexsym}
\usepackage{amsmath}
\usepackage{amssymb}
\usepackage{graphicx}
\usepackage{tikz,tikz-qtree}
\usetikzlibrary{positioning,arrows,shapes}
\usepackage{algorithm}
\usepackage{algorithmic}
\usepackage{multirow}
\usepackage{gb4ea}
\usepackage{textcomp} 
\usepackage{color}

\usepackage{url}

\aclfinalcopy 

\newcommand{\np}[1]{[$_{\scriptsize\mathsf{NP}}$\,#1]}
\newcommand{\vp}[1]{[$_{\scriptsize\mathsf{VP}}$\,#1]}

\newcommand{\positive}[1]{#1$\uparrow$}
\newcommand{\negative}[1]{#1$\downarrow$}
\newcommand{\up}{$^{+}$}
\newcommand{\down}{$^{-}$}

\newcommand{\modified}[1]{\textcolor{black}{#1}}
\newcommand{\final}[1]{\textcolor{black}{#1}}

\title{HELP: A Dataset for Identifying Shortcomings of \\
Neural Models in Monotonicity Reasoning}

\author{
  \parbox{\linewidth}{\centering
   Hitomi Yanaka$^{1,2}$,
   Koji Mineshima$^2$,
   Daisuke Bekki$^2$,\linebreak
   Kentaro Inui$^{1,3}$,
   Satoshi Sekine$^1$,
   Lasha Abzianidze$^4$, and
   Johan Bos$^4$
  }
  \\
   $^1$\mbox{\rm RIKEN,}
   $^2$\mbox{\rm Ochanomizu University,}
   $^3$\mbox{\rm Tohoku University, Japan} \\ 
$^4$\mbox{\rm University of Groningen, Netherlands}
  \\
  \parbox{\linewidth}{\centering
   {\tt \{hitomi.yanaka, satoshi.sekine\}@riken.jp},
   {\tt mineshima.koji@ocha.ac.jp},
   {\tt bekki@is.ocha.ac.jp},
   {\tt inui@ecei.tohoku.ac.jp},
   {\tt \{l.abzianidze, johan.bos\}@rug.nl}
   }
}

\date{}
\begin{document}
\maketitle
\begin{abstract}

Large crowdsourced datasets are widely used for training and evaluating neural models on natural language inference (NLI).
Despite these efforts, neural models have a hard time capturing logical inferences, including those licensed by phrase replacements, so-called monotonicity reasoning.
Since no large
dataset has been developed for monotonicity reasoning,
it is still unclear whether the main obstacle is the size of datasets or the model architectures themselves.
To investigate this issue, we introduce a new dataset, called HELP, for handling entailments with lexical and logical phenomena.
We add it to training data for the state-of-the-art neural models and evaluate them on test sets for monotonicity phenomena.
The results showed that our data augmentation
improved the overall accuracy.
\modified{We also find that the improvement is better on
monotonicity inferences with lexical replacements
than on downward inferences with disjunction and modification.
This suggests that some types of inferences can be improved by our data augmentation while others are immune to it.}

\end{abstract}

\section{Introduction}
\label{sec:intro}
Natural language inference (NLI)
has been proposed as a benchmark task for natural language understanding.
\modified{This task is to determine whether a given statement (premise) semantically entails another statement (hypothesis)~\cite{series/synthesis/2013Dagan}. }
Large crowdsourced datasets such as SNLI~\cite{snli:emnlp2015} and MultiNLI~\cite{DBLP:journals/corr/WilliamsNB17} have been
created from naturally-occurring texts for training and testing neural models on NLI.
\final{Recent reports showed that these crowdsourced datasets contain undesired biases that allow prediction of entailment labels only from hypothesis sentences~\cite{gururangan-EtAl:2018:N18-2,poliak-EtAl:2018:S18-2,DBLP:conf/lrec/Tsuchiya18}. Moreover, these standard datasets come with} the so-called \textbf{upward} monotonicity inferences (see Table \ref{table:dataset}), i.e., inferences from subsets to supersets (\textit{changes in personal values} $\sqsubseteq$ \textit{changes in values}), but they rarely come with \textbf{downward} monotonicity inferences, i.e., inferences from supersets to subsets (\textit{commissioners} $\sqsupseteq$ \textit{female commissioners}). Downward monotonicity inferences are interesting in that they allow to replace a phrase with a more specific one and thus the resulting sentence can become longer, yet the inference is valid.

\begin{table}[!bt]
\scalebox{0.58}{
\begin{tabular}{ll} \hline
\textsf{Upward} &
\textit{Some \textbf{changes in personal values} are simply part of growing older}\\
(MultiNLI) &
$\Rightarrow$\ \textit{Some \textbf{changes in values} are a part of growing older} \\\hline
\textsf{Downward} & 
\textit{At most ten \textbf{commissioners} spend time at home}\\
(FraCaS) &
$\Rightarrow$\ \textit{At most ten \textbf{female commissioners} spend time at home}\\\hline
\end{tabular}
}
\caption{\label{tab:nli} Upward and downward inferences.}
\label{table:dataset}
\end{table}

FraCaS~\cite{cooper1994fracas} 
\modified{contains such logically challenging problems as downward inferences}.
\modified{However, it is small in size (only 346 examples) for training neural models,
and it covers only simple syntactic patterns with severely restricted vocabularies.}
The lack of such a dataset \modified{on a large scale} is due to at least two factors:
it is hard to instruct crowd workers \modified{without deep knowledge of natural language syntax and semantics,}
and it is also unfeasible to employ experts to obtain a large number of logically challenging inferences.

\citet{Bowman2015} proposed an artificial dataset for logical reasoning, whose premise and hypothesis are automatically generated from a simple English-like grammar.
Following this line of work,
\citet{Geiger2018} presented a method to construct a complex dataset for multiple quantifiers \modified{(e.g., \textit{Every dwarf licks no rifle} $\Rightarrow$ \textit{No ugly dwarf licks some rifle})}.
These datasets contain downward inferences, but they are designed not to require lexical knowledge.
There are also NLI datasets 
which expand lexical \modified{knowledge}
by replacing words
using lexical rules~\cite{Monz2001LightWeightEC,glockner-shwartz-goldberg:2018:Short,naik-EtAl:2018:C18-1,poliak-EtAl:2018:BlackboxNLP}.
In these works, however,
little attention has been paid to
downward inferences.

\modified{The GLUE leaderboard~\cite{wang2018glue} reported that neural models did not perform well on downward inferences, and this leaves us guessing whether the lack of large datasets for such kind of inferences that involve the interaction between lexical and logical inferences is an obstacle of understanding inferences for neural models.}

To shed light on this problem, this paper makes the following three contributions:
\modified{(a) providing a method to create a large NLI dataset\footnote{Our dataset and its generation code will be made publicly available at https://github.com/verypluming/HELP.} that embodies the combination of lexical and logical inferences focusing on monotonicity (i.e., phrase replacement-based reasoning) (Section~\ref{sec:lex}),}
(b) measuring to what extent the new dataset helps neural models to learn 
\modified{monotonicity inferences}, and 
(c) by analyzing the results, revealing which types of logical inferences are solved with our training data augmentation and which ones are immune to it (Section~\ref{sec:result}).

\section{Monotonicity Reasoning}
\label{sec:background}
Monotonicity reasoning is a sort of reasoning based on word replacement.
Based on the monotonicity properties of words, it determines whether a certain word replacement results in a sentence entailed from the original one~\cite{10.2307/25001141,moss2014}.
\modified{A polarity is a characteristic of a word position imposed by monotone operators. 
Replacements with more general (or specific) phrases in $\uparrow$ (or $\downarrow$) polarity positions license entailment.
Polarities are determined by a function which is always upward monotone ($+$) (i.e., an order preserving function that licenses entailment from specific to general phrases), always downward monotone ($-$) (i.e., an order reversing function) or neither, non-monotone.}

Determiners are modeled as binary operators, taking noun and verb phrases as the first and second arguments, respectively, and they entail sentences with their arguments narrowed or broadened
according to their monotonicity properties.
For example, the determiner \textit{some} is upward monotone
both in its first and second arguments,
and
the concepts can be broadened by replacing its hypernym (\textit{people} $\sqsupseteq$ \textit{boy}), removing modifiers (\textit{dancing} $\sqsupseteq$ \textit{happily dancing}), or adding disjunction.
The concepts can be narrowed by replacing its hyponym (\textit{schoolboy} $\sqsubseteq$ \textit{boy}), adding modifiers, or adding conjunction.
\\[1ex]
\scalebox{0.80}{
\begin{tabular}{ll}
(1) &\textit{Some} \np{\positive{\textit{boys}}}\up \vp{\positive{\textit{are happily dancing}}}\up\\
$\Rightarrow$& \textit{Some} \np{\textit{people}} \vp{\textit{are dancing}} \\
$\nRightarrow$& \textit{Some} \np{\textit{schoolboys}} \vp{\textit{are dancing and singing}} \\
\end{tabular}
}
\\[1ex]

If \final{a sentence contains negation}, the polarity of words over the scope of negation is reversed:%
\\[1ex]
\scalebox{0.80}{
\begin{tabular}{ll}
(2) &\textit{No} \np{\negative{\textit{boys}}}\down \vp{\negative{\textit{are happily dancing}}}\down\\
$\nRightarrow$&\textit{No} \np{\textit{one}} \vp{\textit{is dancing}}\\
$\Rightarrow$&\textit{No} \np{\textit{schoolboys}} \vp{\textit{are dancing and singing}}\\
\end{tabular}
}
\\[1ex]
\noindent If the propositional object is embedded in another negative or conditional context, the polarity of words over its scope can be reversed again:
\\[1ex]
\scalebox{0.80}{
\begin{tabular}{ll}
(3) &\textit{If} [\textit{there are no} \np{\positive{\textit{boys}}}\down \vp{\positive{\textit{dancing happily}}}\down]\down,\\
&[\textit{the party might be canceled}]\up\\
$\Rightarrow$&\textit{If} [\textit{there is no} \np{\textit{one}} \vp{\textit{dancing}}],\\
&[\textit{the party might be canceled}] \\
\end{tabular}
}
\\[1ex]
\noindent In this way, \modified{the polarity of words is determined by monotonicity operators and syntactic structures.}

\section{Data Creation}\label{sec:lex}
We address three issues when creating the inference problems:
\modified{(a) Detect the monotone operators and their arguments;}
(b) Based on the syntactic structure, induce the polarity of the argument positions;
(c) Using lexical knowledge or logical connectives, narrow or broaden the arguments.

\subsection{Source corpus}
\label{ssec:source_corpus}

We use sentences from the Parallel Meaning Bank \cite[PMB,][]{abzianidze-EtAl:2017:EACLshort} 
as a source while creating the inference dataset. 
The reason behind choosing the PMB is threefold.
First, the fine-grained annotations in the PMB facilitate our automatic monotonicity-driven construction of inference problems.
In particular, semantic tokenization and WordNet~\cite{wordnet} senses make narrow and broad concept substitutions easy
while the syntactic analyses in Combinatory Categorial Grammar~\cite[CCG,][]{Steedman00} format 
and semantic tags \cite{AbzianidzeBos2017IWCS} contribute to monotonicity and polarity detection.
\modified{Second, the PMB contains lexically and syntactically diverse texts from a wide range of genres.}
Third, the gold (silver) documents are fully (partially) manually verified, which control noise in the automated generated dataset.  
To prevent easy inferences, we use the sentences with more than five tokens from 5K gold and 5K silver portions of the PMB.

\begin{figure}
\centering
\scalebox{0.61}{
\begin{tikzpicture}[node distance=2cm, align = center]
\tikzstyle{inout} = [rectangle, minimum width=2cm, minimum height=0.7cm, draw=black, very thick, rectangle split, rectangle split parts=2, rectangle split part fill={gray!10!white,white}]
\tikzset{arrow/.style={draw, -latex'}}

\node (select) [inout, text width=30em] {
\textbf{Step 1. Select a sentence using semantic tags from the PMB} 
\nodepart{two}
 \begin{tabular}{lllllll}
 \textit{\textbf{All}} & \textit{kids} & \textit{were} & \textit{dancing} & \textit{on} & \textit{the} & \textit{floor} \\
 \textbf{AND} & \texttt{CON} & \texttt{PST} & \texttt{EXG} & \texttt{REL} & \texttt{DEF} & \texttt{CON} 
 \end{tabular}
 };
\node (ccg) [inout, text width=30em, below = 0.5cm of select] {
\textbf{Step 2. Detect the polarity of constituents via CCG analysis}
\nodepart{two}
\textit{All} \np{\negative{\textit{kids}}} were \vp{\positive{\textit{dancing on the floor}}}
};

\node (word) [inout, text width=30em, below = 0.5cm of ccg] {
\textbf{Step 3. Replace expressions based on monotonicity}
\nodepart{two}
\begin{tabular}{rl}
$P$:& \textit{All} \np{\negative{\textit{kids}}} \vp{\positive{\textit{were dancing on the floor}}}\\
$H_1$:& \textit{All} \np{\textit{\textbf{foster children}}} \vp{\textit{were dancing on the floor}} 
\hfill{\underline{\textsc{entail}}}\\
$H_2$:& \textit{All \np{kids}} \vp{\textit{\textbf{were dancing}}} \hfill{\underline{\textsc{entail}}}
\end{tabular}};
\node (swap) [inout, text width=30em, below = 0.5cm of word] {
\textbf{Step 4. Create another inference pair by swapping sentences}
\nodepart{two}
\begin{tabular}{rl}
$P_1' (= H_1)$:& \textit{All} \np{\textit{foster children}} \vp{\textit{were dancing on the floor}}\\
$P_2' (= H_2)$:& \textit{All \np{kids}} \vp{\textit{were dancing}}\\
$H' (= P)$:& \textit{All} \np{\textit{kids}}  \vp{\textit{were dancing on the floor}} \hfill{\underline{\textsc{neutral}}}
\end{tabular}};

\draw[arrow] (select) edge (ccg);
\draw[arrow] (ccg) edge (word);
\draw[arrow] (word) edge (swap);
\end{tikzpicture}
}
\vspace{-0.25cm}
\caption{Illustration for creating the HELP dataset.
}
\label{pipeline}
\vspace{-0.25cm}
\end{figure}
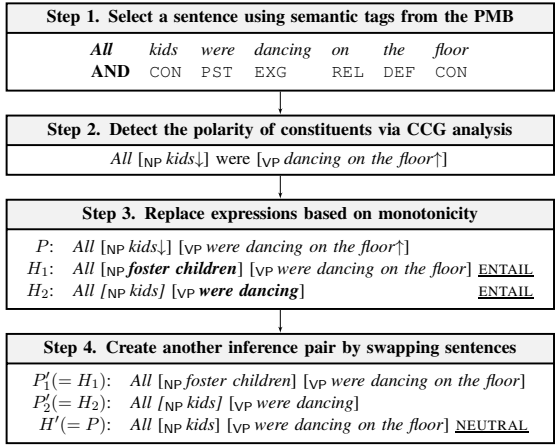

\subsection{Methodology}
Figure~\ref{pipeline} illustrates the method of creating
the HELP dataset.
We use declarative sentences from the PMB containing 
\modified{monotone operators, conjunction, or disjunction} as a source (Step 1).
These target words can be identified by their semantic tags:
\texttt{AND} (\textit{all}, \textit{every}, \textit{each}, \textit{and}),
\texttt{DIS} (\textit{some}, \textit{several}, \textit{or}), \modified{\texttt{NEG} (\textit{no}, \textit{not}, \textit{neither}, \textit{without})}, \texttt{DEF} (\textit{both}),
\texttt{QUV} (\textit{many}, \textit{few}),
\modified{and \texttt{IMP} (\textit{if}, \textit{when}, \textit{unless})}.
In Step 2, after locating the first ({\small\textsf{NP}}) and the second  ({\small\textsf{VP}}) arguments of the
\modified{monotone operator}
via a CCG derivation, 
we detect their polarities with the possibility of reversing a polarity if an argument appears in a \modified{downward environment}.

\modified{In Step 3, to broaden or narrow the first and the second arguments,
we consider two types of operations: (i) lexical replacement, i.e., substituting the argument with its hypernym/hyponym (e.g., $H_1$) and (ii) syntactic elimination, i.e., dropping a modifier or a conjunction/disjunction phrase in the argument (e.g., $H_2$).
Given the polarity of the argument position ($\uparrow$ or $\downarrow$) and the type of replacement (with more general or specific phrases), the gold label (\textit{entailment} or \textit{neutral}) of a premise-hypothesis pair
is automatically determined; e.g., both $(P, H_1)$ and $(P, H_2)$ in Step 3 are assigned \textit{entailment}}.
\modified{For
lexical replacements, we use WordNet senses from the PMB and their ISA relations with the same part-of-speech to control naturalness of the obtained sentence.}
To compensate missing word senses from the silver documents, we use the Lesk algorithm~\cite{Lesk:1986:ASD:318723.318728}.
\modified{In Step 4, by swapping the premise and the hypothesis, we create another inference pair and assign its gold label; e.g., $(P_1', H')$ and $(P_2', H')$ are created and assigned \textit{neutral}.
By swapping a sentence pair created by syntactic elimination, we can create a pair such as $(P_2', H')$ in which the hypothesis is more specific than the premise.
}

\subsection{The HELP dataset}
The resulting dataset has 36K inference pairs consisting of \modified{upward monotone, downward monotone, non-monotone, conjunction, and disjunction}.
Table~\ref{tab:ex} shows some examples.
The number of vocabulary items is 15K.
\modified{
We manually checked the naturalness of randomly sampled 500 sentence pairs, of which 146 pairs were unnatural.
\final{As mentioned in previous work~\cite{glockner-shwartz-goldberg:2018:Short}}, there are some cases where WordNet for substitution leads to unnatural sentences due to the context mismatch;
\final{e.g., an example such as \textit{P}: \textit{You have no driving happening} $\Rightarrow$ \textit{H}: \textit{You have no driving experience}, where \textit{P} is obtained
from \textit{H} by replacing \textit{experience} by
its hypernym \textit{happening}.}
Since our intention is to explore possible ways to augment training data for monotonicity reasoning, we include these cases in the training dataset.}

\begin{table}[!bt]
\begin{center}
\scalebox{0.70}{
\begin{tabular}{lll} \hline
                         Section & Size & Example \\ \hline \hline
   \multirow{2}{*}{Up}& \multirow{2}{*}{7784} & \textit{Tom bought some \textbf{Mexican sunflowers} for Mary} \\
   &&$\Rightarrow$\textit{Tom bought some \textbf{flowers} for Mary*}  \\ \hline
   \multirow{2}{*}{Down} & \multirow{2}{*}{21192} &\textit{If there's no \textbf{water}, there's no whisky*} \\
   &&$\Rightarrow$\textit{If there's no \textbf{facility}, there's no whisky} \\ \hline
   \multirow{2}{*}{Non} & \multirow{2}{*}{1105} &\textit{Shakespeare wrote both \textbf{tragedy and comedy}*} \\
   &&$\nRightarrow$\textit{Shakespeare wrote both \textbf{tragedy and drama}} \\ \hline
   \multirow{2}{*}{Conj}& \multirow{2}{*}{6076} &  \textit{Tom removed his glasses} \\
   &&$\nRightarrow$\textit{Tom removed his glasses \textbf{and rubbed his eyes}*} \\ \hline
   \multirow{2}{*}{Disj}& \multirow{2}{*}{438} & \textit{The trees are barren} \\
   &&$\Rightarrow$\textit{The trees are barren \textbf{or bear only small fruit}*}  \\ \hline
\end{tabular}
}
\vspace{-0.25cm}
\caption{\label{tab:ex} Examples in HELP. The sentence with an asterisk is the original sentence from the PMB.}
\end{center}
\vspace{-0.4cm}
\end{table}

\section{Experiments}
We use HELP as additional training material for three neural models for NLI and evaluate them on
test sets dealing with monotonicity reasoning.

\begin{table*}[!hbt]
\begin{center}
\scalebox{0.59}{
\begin{tabular}{l|l|rr|rr|rr|rr|rr|rr|rr|rr|rrrr} \hline
                         \multirow{4}{*}{Model} && 
                        \multicolumn{12}{c|}{GLUE diagnostic
                        } & \multicolumn{2}{c|}{FraCaS} & \multicolumn{2}{c|}{SICK} &
                        \multicolumn{4}{c}{MNLI}\\
    &Train&\multicolumn{2}{c|}{Up}&\multicolumn{2}{c|}{Down}&\multicolumn{2}{c|}{Non}&\multicolumn{2}{c|}{Conj}&\multicolumn{2}{c|}{Disj}&\multicolumn{2}{c|}{Total}&&&&&\multicolumn{2}{c}{match}&\multicolumn{2}{c}{mismatch}\\ 
    &Data&\multicolumn{2}{c|}{(30)}&\multicolumn{2}{c|}{(30)}&\multicolumn{2}{c|}{(22)}&\multicolumn{2}{c|}{(32)}&\multicolumn{2}{c|}{(38)}&\multicolumn{2}{c|}{(152)}&\multicolumn{2}{c|}{(80)}&\multicolumn{2}{c|}{(4927)}&\multicolumn{2}{c}{(10000)}&\multicolumn{2}{c}{(10000)}\\ 
    &&&$\triangle$&&$\triangle$&&$\triangle$&&$\triangle$&&$\triangle$&&$\triangle$&&$\triangle$&&$\triangle$&&$\triangle$&&$\triangle$\\ \hline \hline
  \multirow{3}{*}{BERT} & MNLI 
  & 50.4 && -67.5 && 23.1 && 52.5 && -6.1 &&17.8&&  65.0 && 55.4&&\textbf{84.6}&&\textbf{83.4}&\\ 
    & +MQ 
    &59.6&\textcolor{blue}{+9.2}&-49.3&\textcolor{blue}{+18.2}&14.0&\textcolor{blue}{-9.1}&62.1&\textcolor{blue}{+9.6}&-18.8 &\textcolor{blue}{-12.7}& 26.3 &\textcolor{blue}{+8.5}& 
    \textbf{68.8} &\textcolor{blue}{+3.8}& 58.2 &\textcolor{blue}{+2.8}&78.4 &\textcolor{blue}{-6.2}&78.6&\textcolor{blue}{-4.8}\\ 
    & +HELP 
    & \textbf{67.0} &\textcolor{blue}{+16.6}& \textbf{29.8}&\textcolor{blue}{+97.3}& \textbf{47.9} &\textcolor{blue}{+24.8}& \textbf{72.1} &\textcolor{blue}{+19.6}& \textbf{-4.1} &\textcolor{blue}{+2.0}&\textbf{51.2} &\textcolor{blue}{+33.4}& 
    \textbf{68.8} &\textcolor{blue}{+3.8}& \textbf{60.0} &\textcolor{blue}{+4.6}&84.4&\textcolor{blue}{-0.2}&83.1 &\textcolor{blue}{-0.3}\\ \hline

  BiLSTM & MNLI 
  & 22.2 && -9.4 && -2.7 && 42.4 && -9.9 && -3.5 && 
  68.9 && 53.8 && \textbf{76.4} && \textbf{76.1} &\\
    +ELMo & +MQ 
    & 22.2 & \textcolor{blue}{0.0} & 8.1 & \textcolor{blue}{+17.5} & -5.7 & \textcolor{blue}{-3.0} & 42.4 & \textcolor{blue}{0.0} & \textbf{-9.8} & \textcolor{blue}{+0.1} & 5.7 & \textcolor{blue}{+9.2} & 
    65.9 & \textcolor{blue}{-3.0} & \textbf{54.0} & \textcolor{blue}{+0.2} & 71.4 & \textcolor{blue}{-5.0} & 70.7 & \textcolor{blue}{-5.4}\\
  +Attn & +HELP 
  & \textbf{32.4} & \textcolor{blue}{+10.2} & \textbf{22.9} & \textcolor{blue}{+32.3} & \textbf{3.7} & \textcolor{blue}{+6.4} & \textbf{45.6} & \textcolor{blue}{+3.2} &  -9.9 & \textcolor{blue}{0.0} & \textbf{17.0} & \textcolor{blue}{+20.5} & 
  \textbf{71.3} & \textcolor{blue}{+2.4} & \textbf{54.0} & \textcolor{blue}{+0.2} & 75.2 & \textcolor{blue}{-1.2} & 74.1 & \textcolor{blue}{-2.0}\\ \hline

  \multirow{3}{*}{ESIM} & MNLI 
  & 14.9 && -14.0 && 6.0 && 29.8 && -3.6 && 1.1 && 
  47.5 && 43.9 && \textbf{71.3} && \textbf{70.7} &\\
    & +MQ 
    & 27.2 & \textcolor{blue}{+12.3} & -7.8 & \textcolor{blue}{+6.2} & 3.4 & \textcolor{blue}{-2.6} & 5.2 & \textcolor{blue}{-24.6} & -13.0 & \textcolor{blue}{-9.4} & 6.8 & \textcolor{blue}{+5.7} & 
    43.7 & \textcolor{blue}{-3.8} & 53.1 & \textcolor{blue}{+9.2} & 68.6 & \textcolor{blue}{-3.7} & 68.2 & \textcolor{blue}{-2.5} \\
    & +HELP 
    & \textbf{31.4} & \textcolor{blue}{+16.5} & \textbf{24.7} & \textcolor{blue}{+38.7} & \textbf{8.0} & \textcolor{blue}{+2.0} & \textbf{32.6} & \textcolor{blue}{+2.8} & \textbf{7.1} & \textcolor{blue}{+10.7} & \textbf{27.0} & \textcolor{blue}{+25.9} & 
    \textbf{48.8} & \textcolor{blue}{+1.3} & \textbf{56.6} & \textcolor{blue}{+12.7} & 71.1 & \textcolor{blue}{-0.2} & 70.1 & \textcolor{blue}{-0.6} \\ \hline

\end{tabular}
}
\vspace{-0.25cm}
\caption{\label{tab:eval} Evaluation results on the GLUE diagnostic dataset, FraCaS, SICK, and MultiNLI (MNLI). The number in parentheses is the number of problems in each test set. $\triangle$ is the difference from the model trained on MNLI.}
\end{center}
\vspace{-0.5cm}
\end{table*}

\subsection{Experimental settings}

\paragraph{Models}
We used three models: BERT~\cite{bert2018}, 
BiLSTM+ELMo+Attn~\cite{wang2018glue}, and
ESIM~\cite{chen-EtAl:2017:Long3}.

\vspace{-1ex}

\paragraph{Training data}
We used three different training sets and compared their performance; MultiNLI (392K), MultiNLI+MQ (the dataset for multiple quantifiers introduced in Section~\ref{sec:intro};~\citeauthor{Geiger2018}, \citeyear{Geiger2018}) (892K),
and MultiNLI+HELP (429K). 

\vspace{-1ex}

\paragraph{Test data}
We used four 
test sets:
(i) the GLUE diagnostic dataset~\cite{wang2018glue} (upward monotone, downward monotone, non-monotone, conjunction, and disjunction sections), (ii) FraCaS (the generalized quantifier section), (iii) the SICK~\cite{MARELLI14.363} test set, \modified{and (iv) MultiNLI matched/mismatched test set}.
We used the Matthews correlation coefficient (ranging $[-1,1]$) as the evaluation metric for GLUE.
Regarding \modified{other datasets}, we used accuracy as the metric.
\modified{We 
also check if our data augmentation does not decrease the performance on MultiNLI.}

\subsection{Results and discussion}
\label{sec:result}
Table~\ref{tab:eval} shows that
adding HELP to MultiNLI improved the accuracy of all models on 
\modified{GLUE, FraCaS, and SICK}.
\modified{Regarding MultiNLI,
note that adding data for downward inference can be harmful for performing upward inference, because lexical replacements work in an opposite way in downward environments.
However, 
our data augmentation minimized the decrease in performance on MultiNLI.
This suggests that models managed to learn the relationships between downward operators and their arguments from HELP.}

The improvement in accuracy is better with HELP than that with MQ \modified{despite the fact that} the size of HELP is much smaller than MQ.
MQ does not
\modified{deal with lexical replacements}, and thus the improvement is not stable.
This indicates that the improvement
comes from carefully controlling the target reasoning of the training set rather than from its size.
ESIM showed a greater improvement in accuracy compared with the other models when we added HELP.
This result arguably supports the finding in \citet{Bowman2015} that a tree architecture is better for learning some logical inferences.
Regarding the evaluation on SICK, \citet{stergios2018} reported a drop in accuracy of 40-50\% when BiLSTM and ESIM were trained on MultiNLI because SICK is out of the domain of MultiNLI.
Indeed, the accuracy of each model, including BERT, was low at 40-60\%.

When compared among linguistic phenomena, the improvement by adding HELP was better for \modified{upward and downward monotone.}
In particular, all models except models trained with HELP failed to answer 68 problems for monotonicity inferences
\modified{with lexical replacements.}
This indicates that such inferences can be improved by adding HELP.

\modified{The improvement for disjunction was smaller than other phenomena. To investigate this, we conducted error analysis on 68 problems of GLUE and FraCaS, which all the models misclassified.}
\modified{44} problems are neutral problems in which all words in the hypothesis occur in the premise (e.g., \textit{He is either in London or in Paris} 
$\nRightarrow$
\textit{He is in London}).
\modified{13} problems are entailment problems in which the hypothesis contains a word or a phrase not occurring in the premise
(e.g., \textit{I don't want to have to keep entertaining people}
$\Rightarrow$
\textit{I don't want to have to keep entertaining people who don't value my time}).
These problems contain disjunction or \modified{modifiers in downward environments where either (i) the premise $P$ contains all words in the hypothesis $H$ yet the inference is invalid or (ii) $H$ contains more words than those in $P$ yet the inference is valid.}\footnote{Interestingly, certain logical inferences including disjunction and downward monotonicity are difficult also for humans to get~\cite{doi:10.1093/jos/ffh018}.}
Although HELP 
contains 21K such
problems, the models nevertheless misclassified them.
This indicates that the difficulty in learning \modified{these non-lexical downward inferences} might not come from the lack of training datasets.

\section{Conclusion and Future Work}
\label{sec:conc}
We introduced
a monotonicity-driven NLI
\modified{data augmentation method.}
The experiments showed that neural models trained on HELP obtained the higher overall accuracy.
\modified{
However, 
the improvement tended to be small 
on downward monotone inferences with disjunction and modification,
which suggests that some types of inferences
can be improved by adding data while others might require different kind of \emph{help}.}

For future work, 
\modified{our
data augmentation} 
can be used for multilingual corpora.
Since the PMB annotations sufficed for creating HELP, applying our method to the non-English PMB documents seems straightforward.  
Additionally, it is interesting to verify the quality and contribution of
a dataset which will be created by using our method on an automatically annotated and parsed corpus. 

\section*{Acknowledgement}
We thank our three anonymous reviewers for helpful suggestions.

\bibliographystyle{acl_natbib}
\bibliography{naaclhlt2019}

\end{document}